\documentclass[lettersize,journal]{IEEEtran}
\usepackage{amsmath,amsfonts}
\usepackage{algpseudocode}
\usepackage{algorithm}
\usepackage{array}
\usepackage[caption=false,font=normalsize,labelfont=sf,textfont=sf]{subfig}
\usepackage{textcomp}
\usepackage{stfloats}
\usepackage{url}
\usepackage{verbatim}
\usepackage{graphicx}
\usepackage{cite}
\usepackage{footnote}
\usepackage[bottom]{footmisc}
\usepackage{threeparttable}
\usepackage{multirow}
\usepackage{tabularx}
\usepackage{ragged2e}
\usepackage{booktabs}
\usepackage{makecell}
\usepackage{xspace}
\usepackage{diagbox}
\usepackage{xcolor}
\usepackage{soul}
\usepackage[final]{pdfpages}
\usepackage{hyperref}

\usepackage{amsthm}
\newtheorem*{assumption*}{\assumptionnumber}
\providecommand{\assumptionnumber}{}
\makeatletter
\newenvironment{assumption}[2]
 {%
  \renewcommand{\assumptionnumber}{Assumption #1.~({#2})}%
  \begin{assumption*}%
  \protected@edef\@currentlabel{#1}%
 }
 {%
  \end{assumption*}
 }
 
\newtheorem*{theorem*}{\theoremnumber}
\providecommand{\theoremnumber}{}
\makeatletter
\newenvironment{theorem}[2]
 {%
  \renewcommand{\theoremnumber}{Theorem #1.~({#2})}%
  \begin{theorem*}%
  \protected@edef\@currentlabel{#1}%
 }
 {%
  \end{theorem*}
 }
 
\makeatother

\newcolumntype{C}[1]{>{\centering\arraybackslash}p{#1}}
\newcolumntype{P}[1]{>{\centering\arraybackslash}p{#1}}
\newcolumntype{M}[1]{>{\centering\arraybackslash}m{#1}}
\newcolumntype{L}[1]{>{\raggedright\arraybackslash}p{#1}}
\newcolumntype{R}[1]{>{\raggedleft\arraybackslash}p{#1}}
\newcolumntype{J}[1]{>{\justifying\arraybackslash}p{#1}}

\newcommand{\myMethod}{\mbox{FedCaP}\xspace} 
\newcommand{\mypara}[1]{\noindent\textbf{#1}}
\newcommand{\Crs}[1]{{\color{black}{#1}}} 

\begin{document}
\title{FedBEVT: Federated Learning Bird's Eye View Perception  Transformer in Road Traffic Systems}

\author{
    Rui Song\textsuperscript{\rm *},
    Runsheng Xu\textsuperscript{\rm *},
    Andreas Festag,
    Jiaqi Ma, and
    Alois Knoll,
\thanks{* The first two authors contributed equally.}
\thanks{This work was supported by the German Federal Ministry for Digital and Transport (BMVI) in the projects ``KIVI -- KI im Verkehr Ingolstadt'' and ''5GoIng – 5G Innovation Concept Ingolstadt''.}%
\thanks{Rui~Song and Andreas~Festag are with Fraunhofer Institute for Transportation and Infrastructure Systems IVI, Ingolstadt, 85051, Germany, e-mail:
        {\tt\small \{rui.song, andreas.festag\}@ivi.fraunhofer.de}.}%
\thanks{Rui~Song and Alois~Knoll are with the Chair of Robotics, Artificial Intelligence and Real-Time Systems, Technical University of Munich, Garching, 85748, Germany, e-mail:
        {\tt\small rui.song@tum.de, knoll@in.tum.de}.}%
\thanks{Runsheng~Xu and Jiaqi~Ma with the Department of Civil and Environmental Engineering,
University of California, Los Angeles, CA 90024 USA, e-mail:
        {\tt\small rxx3386@ucla.
edu, majiaqimark@gmail.com}.}%
\thanks{Andreas~Festag is with Technische Hochschule Ingolstadt, CARISSMA Institute for Electric, COnnected, and Secure Mobility (C-ECOS), Ingolstadt, 85049, Germany, e-mail:
        {\tt\small andreas.festag@carissma.eu}.}%
}



\maketitle

\begin{abstract}
Bird's eye view (BEV) perception is becoming increasingly important in the field of autonomous driving. It uses multi-view camera data to learn a transformer model that directly projects the perception of the road environment onto the BEV perspective. However, training a transformer model often requires a large amount of data, and as camera data for road traffic are often private, they are typically not shared. Federated learning offers a solution that enables clients to collaborate and train models without exchanging data but model parameters. 
In this paper, we introduce FedBEVT, a federated transformer learning approach for BEV perception. \Crs{In order to address two common data heterogeneity issues in FedBEVT: (i) diverse sensor poses, and (ii)~varying sensor numbers in perception systems, we propose two approaches --  Federated Learning with Camera-Attentive Personalization~(FedCaP) and Adaptive Multi-Camera Masking~(AMCM), respectively.}
To evaluate our method in real-world settings, we create a dataset consisting of four typical federated use cases. Our findings suggest that FedBEVT outperforms the baseline approaches in all four use cases, demonstrating the potential of our approach for improving BEV perception in autonomous driving. 
\end{abstract}

\section{Introduction}
\label{sec:intro}

Recently, there has been a significant surge of interest in the bird's-eye-view (BEV) perception for autonomous driving. BEV representations of traffic scenarios are particularly appealing for several reasons. Firstly, as all autonomous vehicles (AVs) are located at ground level \cite{xia2023automated,meng2023hydro}, the BEV representation can omit the z-axis to make the perception results more efficient~\cite{xu2022cobevt}. %
Secondly, it provides rich context and geometry information that can be directly utilized for downstream tasks such as planning~\cite{ma2022vision,liu2023systematic}. Especially, within the first comprehensive survey addressing control systems for autonomous vehicles and connected and automated vehicles \cite{liu2023systematic}, the authors thoroughly expound on the pivotal role of perception systems in enabling low-level control. Finally, the BEV representation can be a unified space for data from different sensor modalities (e.g., camera, LiDAR) and timestamps to fuse without much extra effort~\cite{liu2022bevfusion}. Traditionally, achieving temporal alignment in multi-source sensor fusion requires additional robust algorithms, such as the popular estimation-prediction integrated framework proposed in \cite{liu2021automated} and the computation light multi-sensor fusion localization algorithm in GPS challenging scenarios\cite{xia2023integrated}.

\begin{figure}[t]
   \centering
   \includegraphics[width=0.5\textwidth]{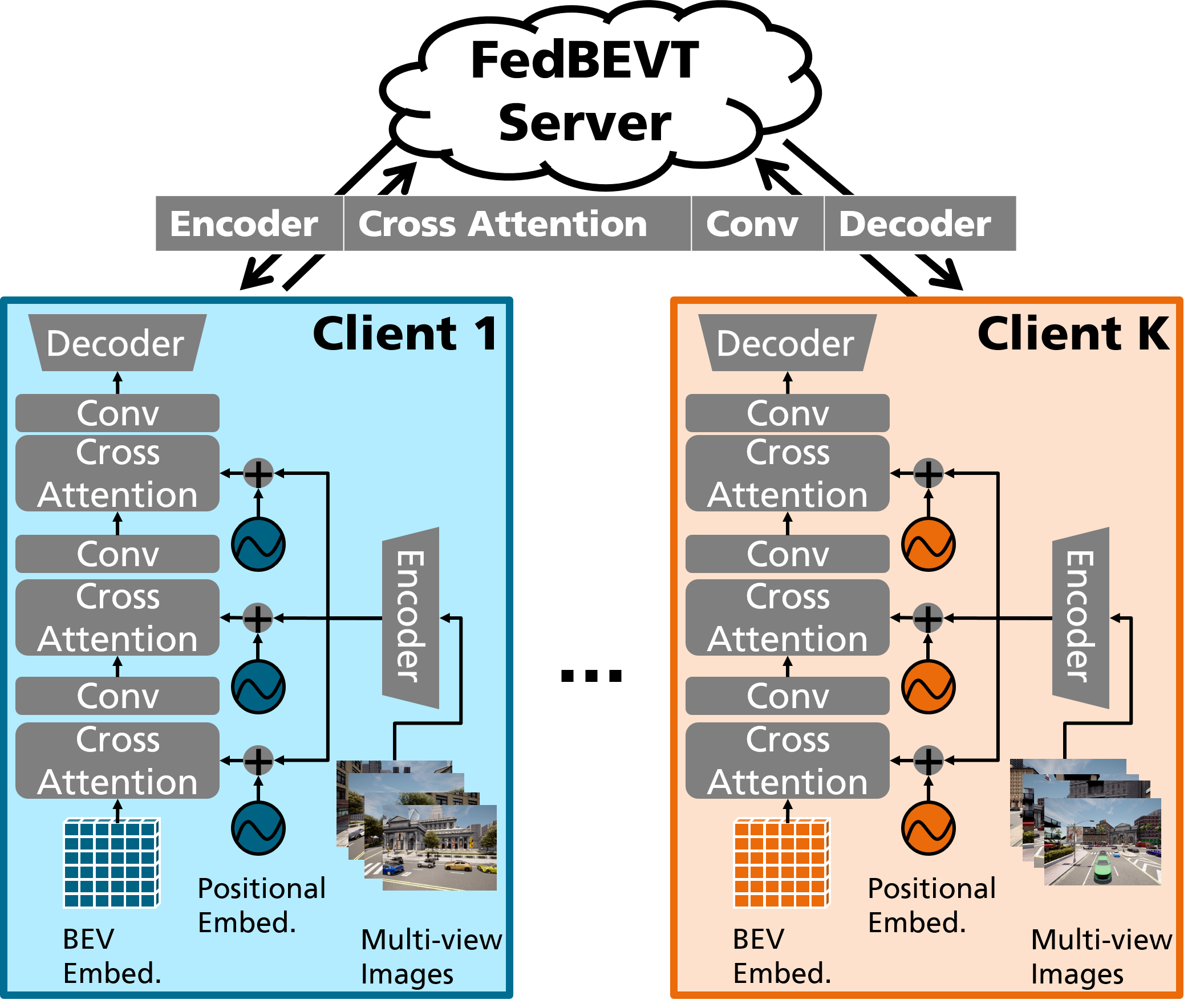}
   \caption{FedBEVT with camera-attentive personlization. The positional embeddings are considered as private parameters for each client. Other parts of BEVT (in gray) are shared to the server for an aggregation.}
   \label{fig:system}
\end{figure}

BEV perception research has begun exploring the use of multi-view camera data for predicting BEV maps due to their low cost~\cite{li2022bevformer,huang2021bevdet,zhou2022cross,liu2022petr,swerdlow2023street}. Nevertheless, inferring 3D information from 2D data is challenging, as mono cameras only provide 2D information. Recent works have thus utilized vision transformers, known for their ability to reason about correlations between different data, to solve the 2D-to-3D problem~\cite{li2022bevformer, fiery2021, zhou2022cross, liu2022petrv2, zhang2022beverse, xu2022cobevt}. Despite promising results, these transformer-based methods have been trained on limited public datasets, such as NuScenes~\cite{caesar2020nuscenes}, which may not generalize well. Companies such as automotive original equipment manufacturers (OEMs), sensor suppliers, software solution providers, and research institutes, need to install multi-sensor systems on several vehicles and gather large amounts of data during extended driving sessions that can be used as the training dataset. %
Nevertheless, such data is often private and can be prohibitively expensive, with some being completely unshared. 
In this context, the advent of federated learning provides a solution that enables collaborative training processes without data exchange and addresses the issue of data privacy.

The federated learning approach offers many benefits but can lead to data heterogeneity while training across clients. In the context of camera-based BEV perception, this challenge of data heterogeneity is caused by the wide range of sensor configurations, with variations in the number of cameras and poses of installed sensors on vehicles or transportation infrastructure.~\cite{philion2020lift}.
These variations can result in significant differences in data characteristics across clients, which vanilla federated learning methods may not be able to overcome. In this paper, we aim to address the practical question of how to leverage the advantages of external data while minimizing the impact of data heterogeneity in a federated learning framework to achieve the highest possible performance improvement for local models.

To address the above data heterogeneity challenge, we propose a new federated learning framework for BEV perception transformers (BEVT), called FedBEVT (as shown in Fig.~\ref{fig:system}) that enables selective sharing of specific parts of the BEV transformer with the server. Following the approach of most transformer-based BEV perception works~\cite{zhou2022cross, xu2022cobevt, liu2022petr}, \Crs{we split a BEV vision model into five components}: encoder for image feature extraction, positional embedding that contains camera geometry information, cross-attention module to project front views to bird's-eye view, convolution layers to refine the features in the transformers, and decoder to transfer the BEV representation to the final prediction.

Given the variability of camera poses across different clients, sharing the entire model for global aggregation results in unsatisfactory performance. To address this, we adopt camera-attentive personalization in FedBEVT by privatizing the positional embedding that contains camera intrinsic and extrinsic information. Additionally, to address the varying number of cameras in each client, we introduce adaptive multi-camera masking (AMCM) to train multi-view data with a consistent BEV size by overlapping masks, which are built based on the combined field of views of all cameras. This ensures a consistent BEV embedding size during federated learning and enables clients with varying camera systems to learn together. To evaluate the effectiveness of our methods, we create a dataset with these variations and distribute it among various clients in federated settings, simulating real-world federated learning scenarios in intelligent traffic systems. Our experiments demonstrate that our method achieves better test accuracy with reduced communication costs for most clients.

\noindent
Our contributions can be summarized as followings:
\begin{itemize}
    \item 
    We present a federated learning framework for the BEV transformer in road traffic perception applications. To the best of our knowledge, this is the first federated transformer training framework specifically designed for the BEV perception task.
    \item 
    We provide a benchmark multi-view camera dataset for BEV perception in road traffic scenarios under federated settings. To address the challenge of data heterogeneity, we consider two popular data variations in federated learning application scenarios in C-ITS domain: (\emph{i}) diverse sensor poses and (\emph{ii}) varying sensor numbers in perception systems.
    \item 
    We propose a federated learning framework based on camera-attentive personalization (\myMethod) for training BEV transformer models. By considering the relationship between training data and the transformer, we adjust the personalization aspect of the transformer in federated learning, providing each client with a customized model that better fits their local data, thereby improving the accuracy of local BEV perception.
    \item 
    We introduce an adaptive multi-camera masking method that enables clients with varying numbers of cameras to train models using federated learning.
    \item 
    We formulate four typical use cases in road traffic systems and divide benchmark data into different clients in federated settings accordingly. These distributed datasets are used to demonstrate the performance of training BEVT through our federated learning system and baselines. Our open-source implementation of FedBEVT is publicly available in GitHub~\footnote{\url{https://github.com/rruisong/FedBEVT}}.
\end{itemize}

\section{Related Work}
\label{sec:related_work}

\subsection{BEV Perception in Road Traffic Systems}
BEV perception involves transforming input image sequences from a perspective view to a BEV, enabling perception tasks such as 3D bounding box detection or semantic map segmentation. There are two main approaches to BEV perception: geometric-based and transformer-based methods. Geometric-based methods leverage the natural geometric projection relationship between camera extrinsic and intrinsic parameters to project the perspective view to a BEV. For example, LSS~\cite{philion2020lift} uses a categorical distribution over depth and a context vector to lift 2D images to a frustum-shaped point cloud, which is then splatted onto the BEV plane using the camera extrinsics and intrinsics. BEVDet~\cite{huang2021bevdet} follows a similar framework, with the addition of a BEV encoder to further refine the projected BEV representations. Transformer-based methods, on the other hand, can implicitly utilize the camera geometry information with learnable embeddings. BEVFormer~\cite{li2022bevformer} uses the geometry information to obtain the initialized sampling offset and applies deformable attention to query the image features into the learnable BEV embedding. Although it achieves high accuracy, it is slow. CVT~\cite{zhou2022cross} develops positional embeddings for each individual camera depending on their intrinsic and extrinsic calibrations. These embeddings are added to the image features and learnable BEV embeddings, and a vanilla cross-view attention transformer is applied to fetch the image features to the BEV embedding. Building upon CVT, CoBEVT~\cite{xu2022cobevt} replaces the vanilla attention with a novel fused axial attention to largely save computation cost and construct a hierarchical structure to explore the multi-scale features. As CoBEVT achieves a good trade-off between inference speed and accuracy, this paper selects it as the deployed model for federated learning study.

\Crs{
\subsection{Federated Learning for Intelligent Traffic Systems}
Federated learning brings great opportunities and potential to the intelligent traffic systems. It allows the use of more data to train better-performing models while respecting privacy regulations~\cite{zhang2021real, elbir2022federated, nguyen2022deep, li2021privacy}. Previous research has commenced the exploration of using federated learning to train models pertinent to applications in intelligent transportation systems, such as 2D object recognition and detection~\cite{jallepalli2021federated, xie2022efficient, kong2021federated}, behavior prediction~\cite{aparna2021steering}, localization~\cite{kong2021fedvcp, zhang2023reconfigurable}, and blockchain-based autonomous vehicles~\cite{pokhrel2020federated, zhu2022blockchain, he2021bift, zhang2020blockchain}. Nevertheless, the application of federated learning for BEV perception remains unexplored. Moreover, existing studies~\cite{fantauzzo2022feddrive, song2022federated, du2020federated} primarily address challenges of heterogeneity in federated learning in the context of system diversity and label shift, while the issue of heterogeneous sensor configurations, a primary cause of data heterogeneity, has not been investigated. To bridge these gaps, our work concentrates on the application of BEV perception and strives to address the heterogeneity resulting from sensor configurations.
}

\subsection{Personalized Federated Learning}

Federated learning is a collaborative approach to machine learning that involves a large number of clients working together within a network~\cite{mcmahan2017communication}. 
Typically, a server initiates the learning process and clients download the global model to train it on their local data. The trained local models are then uploaded back to the server, and the global model is updated through aggregation. After several communication rounds, a model that performs well on the global data is obtained. However, such models may not work well with local data, as the local data distribution in most federated learning applications is often different from the global data distribution.

To address the challenge of local data distribution discrepancies in federated learning, personalized federated learning has been proposed as a solution \cite{arivazhagan2019federated, hanzely2021personalized, pillutla2022federated}. This approach customizes the model in each client to account for the unique characteristics of their local data.
 One of the most popular and effective methods for achieving personalized federated learning is the architecture-based approach \cite{tan2022towards}. This method decouples the model's parameters, allowing only a subset of parameters to be shared and aggregated among clients, while the private parameters are learned solely on local data. Previous research has attempted to select these private parameters based on model architecture~\cite{zhang2021parameterized,shamsian2021personalized, liang2020think, yi2023fedgh} or data similarities~\cite{collins2021exploiting,huang2021personalized, bui2019federated}.
However, none of these methods have been proposed specifically for training personalized transformer models, which have a significantly different structure compared to other machine learning models. Therefore, we further study on developing personalized federated learning methods that are tailored to the unique characteristics of transformer models.
    
\subsection{Federated Learning for Transformer}

Transformer is a neural network architecture that uses self-attention mechanisms and was originally designed for natural language processing tasks~\cite{vaswani2017attention}. 
It has also been effective in computer vision tasks such as object recognition (ViT)~\cite{dosovitskiy2021an,xu2022v2x,xu2023v2v4real}. 
As demonstrated in \cite{liu2022yolov5}, the Transformer architecture has displayed a heightened level of performance compared to traditional convolutional neural networks when trained on extensive datasets. Moreover, this architecture showcases promising potential in the domain of detecting small objects, as exemplified in UAV imagery.
Federated learning offers the potential to further enhance the training process by incorporating data from multiple clients, making it a promising approach for training Transformer models. 
Recent experimental results in~\cite{qu2022rethinking} have shown that the Transformer architecture is more robust to distribution shifts and improves federated learning for object detection tasks. 

However, its application to the BEV semantic segmentation task, which is crucial in autonomous driving and intelligent traffic systems, remains an area for further research.
Our observation has revealed that variations in camera setup are the primary source of data heterogeneity in federated learning for BEV semantic segmentation tasks. 
To tackle this challenge, our federated learning framework incorporates personalized positional embedding and integrates cross attention and CNN layers to boost the transformer's performance and leverage larger data amounts.

\section{Methodology}
\label{sec:method}

\begin{figure}[t!]
   \centering
   \includegraphics[width=0.4\textwidth]{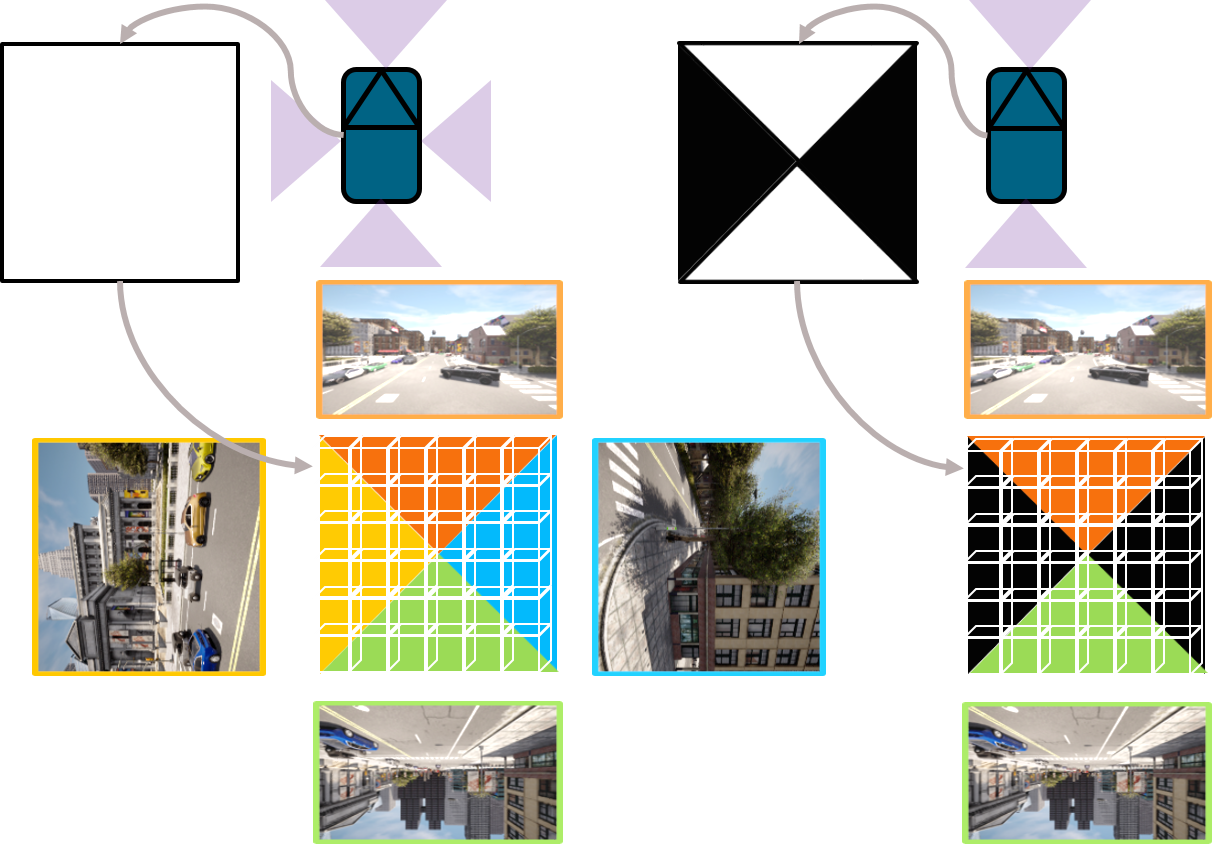}
   \caption{Illustration of AMCM used for two vehicles equipped with different numbers of cameras. By adapting the mask to the total field of view in each perception system, the size of BEV embeddings can be maintained and adjusted for use in FedBEVT.}
   \label{fig:AMCM}
\end{figure}

\begin{figure*}[t!]
   \centering
   \includegraphics[width=1\textwidth]{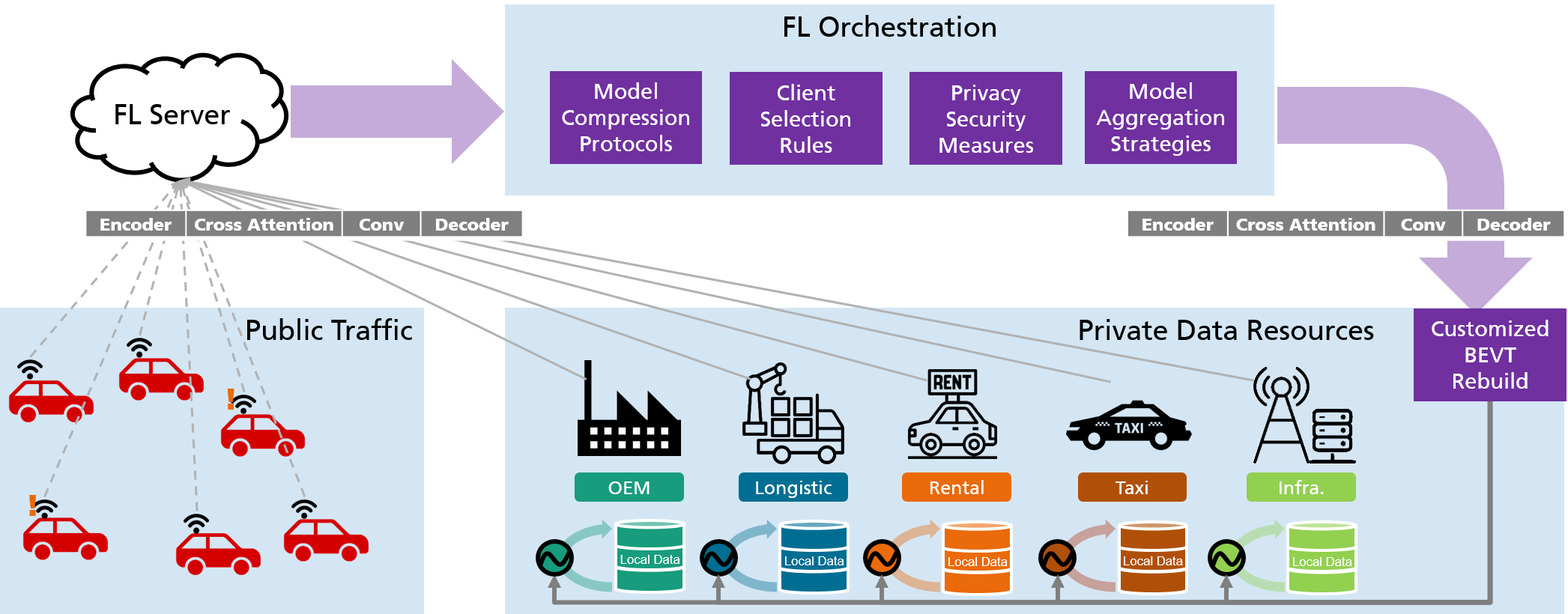}
   \caption{The system overview of FedBEVT illustrates vehicle clients from public traffic and clients from private data resources. The federated learning server manages the training process with protocols for model compression, client selection, and secure aggregation similar to those in a typical federated learning framework for deep learning models. Note that the data heterogeneity exists due to various clients in intelligent road traffic systems.}
   \label{fig:exp}
\end{figure*}

\subsection{Problem Formulation}

In a federated learning training scenario, we consider $K$ clients, each with their own local dataset consisting of $N_k$ data points paired with the corresponding BEV semantic masks $Y_k$. Each data point $X_{k,i}$ contains $L_k$ images from a multi-view camera system, resulting in the format $(X_{k}, Y_{k})$. Here, $X_k \in \mathbb{R}^{N_k \times L_k \times H \times W \times 3}$ and $Y_{k} \in \mathbb{R}^{N_k \times h \times w \times 2}$, where $H$ and $W$ represent the height and width of one image, and $h$ and $w$ represent the height and width of the BEV semantic mask. The semantic mask has two classes, background and vehicles. These clients may come from a variety of industries, including original equipment manufacturers (OEMs), logistics and taxi companies, car rental companies, or infrastructure owners such as traffic management departments.

In general federated learning, each client receives a global model from the server and trains locally for several epochs before returning the trained model to the server for fusion into a new global model. This process continues for a certain number of communication rounds until a trained model is obtained.
 The goal of federated learning is to minimize the following:
\begin{equation}
    \label{Eq:fl}
     f(w) = \frac{1}{N}  \sum_{k=1}^K  \sum_{i=1}^{N_k} f_i (w) = \mathbb{E}_{X_i \in X} [f_i(w)],
\end{equation}
where $f_i(w) = l(BevT(X_i), Y_i; w)$,  $N$ is the number of data points in all clients, $l$ is the loss function, and $BevT$ is a transformer-based BEV model with three major components: image encoder, BEV transformer, and BEV decoder. The image feature encoder utilizes a CNN network to encode multi-camera images. The core of the BEV transformer is usually a multi-layer attention structure~\cite{vaswani2017attention}. In this paper, we employ the Fax-Attention schema~\cite{xu2022cobevt}, which involves both sparse cross-view attention and self-attention between image features and the BEV query. Additionally, BEV query and positional embeddings are required as inputs to the BEVT. The BEV query is a learnable embedding to represent the grid world, and the ultimate goal is to locate the position of other traffic objects in this query. The positional embeddings contain position information of each transformer image feature, including information related to the coordinate system transformation for converting the camera perspective to the BEV perspective. This improves the intuitiveness of the input for generating BEV features. We adopt the approach from~\cite{zhou2022cross} and represent the output of the positional embeddings as follows, assuming the extrinsic parameters of each camera $j$ are $R_{i,j,k} \in \mathbb{R}^{3\times3}$ for rotation and $t_{i,j,k} \in \mathbb{R}^3$:
\begin{equation}
\label{Eq:pos-emb}
z_{i,k} = \mathbf{\Phi}^{(v_k)}(\mathbf{T}(R_{i,j,k}, t_{i,j,k}, p_{i,j,k})_{j=1}^{L_k}).
\end{equation}
Here, the image coordinate-system position features in $p$ are first transformed into the 3D vehicle coordinate-system using $\mathbf{T}$. 
Next, the resulting positions are mapped by $\mathbf{\Phi}^{(v_k)}$, producing a positional embedding $z_{i,k}$. We denote the parameters in $\mathbf{\Phi}$ as ${v_k}$, to distinguish other parameters in the transformer model $u$ used in Sec.~\ref{cap}.

The BEV features are obtained by passing through multiple Fax-Attention layers and processed by a decoder network to produce semantic segmentation outcomes in the BEV perspective, which enables direct determination of the position of surrounding cars on the motion plane. Our approach addresses two key challenges: (\emph{i}) developing tailored models for unique sensor systems, and (\emph{ii}) effectively training models using data from varying numbers of cameras.

\subsection{Camera-Attentive Personalization}
\label{cap}

The variation in camera positions across clients can cause the traditional federated learning results to deviate from the optimal values of local data. This is due to the encoding of camera position differences into positional embeddings during transformer model construction, which directly affects the parameter training.

\Crs{To tackle this challenge, we introduce federated learning with camera-attentive personalization (\myMethod), which involves separating the locally trained parameters $v_k$ in MLP for positional embedding, as shown in Eq.~\ref{Eq:pos-emb}, from the transformer and treating them as private parameters for a given client $k$.
Apart from $v_k$, we share the remaining parameters in $w_k$ in the BEVT and refer to these public parameters as $u_k$ for the client $k$. As a result, the global update in each communication round is represented by the aggregated public parameters from all clients, which we denote as $u$.}
Then the goal of local training is minimizing:
\begin{equation}
    \label{Eq:local}
     F_k(u,v_k) =  \frac{1}{N_k} \sum_{i=1}^{N_k} f_i(u,v_k),
\end{equation}
and the global goal is rewritten from Eq.~\ref{Eq:fl}:
\begin{equation}
    \label{Eq:global}
     f(u,v_1,..,v_K) = \sum_{k=1}^K \frac{N_k}{N} F_k(u, v_k).
\end{equation}
Each client uploads the encoder, decoder, and other transformer parts in BEVT. After the server aggregates these parameters, the global parameters are coupled with the private parameters to train the next round of data. 

\Crs{In particular, The local update for each client, $k$, consists of two stages:
\setcounter{equation}{4}
\begin{align}
\begin{split}
     v_k \leftarrow v_k - \eta_v \nabla_v F(BevT; \mathcal{B}),\\
     u_k \leftarrow u_k - \eta_u \nabla_u F(BevT; \mathcal{B}),
\end{split}
\end{align}
where $\eta_v$ and $\eta_u$ are the learning rates for $v_k$ and $u_k$, respectively\footnote{To consolidate the hyperparameters, we set these learning rates to be equal, such that $\eta = \eta_v = \eta_u$.}, $\mathcal{B}$ is a batch of the local dataset $\mathcal{D}_k$, and the batch size is $\mathbb{B}$. The number of local training epochs is denoted by $E$, and each individual epoch is referenced by the index $e$. 
The shared parameters, $u$, are updated globally through aggregation as follows:
\begin{align}
\begin{split}
\label{eq:agg1}
     u \leftarrow \sum_{k=1}^K \frac{N_k}{N} u_k.
\end{split}
\end{align}
If we consider that only a subset $\mathcal{S}^t$ (where $M=|\mathcal{S}^t|$) of all clients are selected for aggregation, \ref{eq:agg1} can be reformulated as:
\begin{align}
\begin{split}
\label{eq:agg2}
     u \leftarrow \sum_{k\in \mathcal{S}^t} \frac{N_k}{N} u_k.
\end{split}
\end{align}
The loss function for the client $k$ is expressed as:
$F(u,v_k) = \mathbb{E}_{\mathcal{B}\subseteq \mathcal{D}_k}F(u,v_k; \mathcal{B})$.}
\Crs{The pseudocode of \myMethod is outlined in Algorithm~\ref{algo:fedcap}.}

\begin{algorithm}[t]
\Crs{
 \caption{\raggedright \myMethod: Federated Learning with Camera-attentive Personalization}
 \label{algo:fedcap}
 \begin{algorithmic}[1]
 \renewcommand{\algorithmicrequire}{\textbf{Server Input:}}
 \renewcommand{\algorithmicensure}{\textbf{Output:}}
 \newcommand{\algorithmicbreak}{\textbf{break}}
 \newcommand{\BREAK}{\State \algorithmicbreak}
 \Require{number of the communication round $T$}
 \renewcommand{\algorithmicrequire}{\textbf{Client Input:}}
 \Require {learning rate $\eta_v$, $\eta_u$}
 \Require {number of local training epochs $E_k$}
 \Ensure  {$BevT$ with public parameter $u$ and private parameter $v_k$ for each client}
 \State Server initializes $u^0$ to all clients
    \For {communication round $t = 1,2,...,T$}
        \State $\mathcal{S}^t \leftarrow ClientSelection(\mathcal{S})$
        \Comment{optional client selection operation in FL orchestration}
        \For {each client $k \in \mathcal{S}^t$ \textbf{in parallel}}
            \If {client $k$ is selected for the first time}
                \State client $k$ encodes the camera calibration parameters into $v_k$ based on Equation~\ref{Eq:pos-emb}
            \EndIf
            \State $\tilde\Delta u_k \leftarrow ClientUpdate(u)$
            \State $u_k \leftarrow {u} + \tilde\Delta u_k$
        \EndFor
        \State $u \leftarrow Aggregate(\{u_k | \forall k\in \mathcal{S}^t\})$
        \Comment{secure aggregation in the server}
    \EndFor

    \\
    \State \textbf{ClientUpdate}(${u}$)
    \State  $u_k \leftarrow u$
    \State  $BevT \leftarrow Rebuild(u_k, v_k)$
    \For{epoch $e = 1,2,...,E$}
        \For{each batch ${\mathcal{B}\subseteq \mathcal{D}_k}$}
            \State $v_k \leftarrow v_k - \eta_v \nabla_v F(BevT; \mathcal{B})$
            \Comment{update private model weights locally}
            \State $u_k \leftarrow u_k - \eta_u \nabla_u F(BevT; \mathcal{B})$
            \Comment{update public model weights locally}
        \EndFor
    \EndFor
    \State $\Delta u_k \leftarrow u_k - {u}$
    \State $\tilde\Delta u_k \leftarrow Compression(\Delta u_k)$
    \Comment{optional compression operation in FL orchestration}
    \State \textbf{return} $\tilde\Delta u_k$
\end{algorithmic}
}
\end{algorithm}

\subsection{Adaptive Multi-Camera Masking}

For each client, multi-view data points may come from a different number of cameras. To facilitate federated learning across these clients, we introduce adaptive multi-camera masking (AMCM), which adaptively overlays a mask on the BEV query based on the comprehensive Field-of-View (FoV). In essence, all clients initialize the BEV query with a uniform size and spatial dimensions.  

Fig.~\ref{fig:AMCM} demonstrates this process using two clients: the first employs four cameras, while the second uses only two, a front view and a rear view. Based on the total FoV applicable to each client type, the BEV query is masked, thus enabling activation strictly within the FoV area. The relationship between each camera and the BEV query leads to the projection of each client's data points onto a BEV query of consistent size. This facilitates a more effective aggregation process in federated learning.

\subsection{System Framework}

The system overview of our federated learning framework is shown in Fig.~\ref{fig:exp}. The framework considers both public traffic and private data resources as clients. Since connections in wireless networks for CAVs can be unstable, we select clients to avoid the straggler effect during the federated learning process. On the other hand, clients from private data resources are typically customers of the results of federated learning, and their local trained models are always aggregated in all communication rounds due to the high network quality.

To decrease the volume of communication in each round, clients can implement compression techniques, such as sparsification and quantization, prior to communication~\cite{lin2018deep, song2022resfed}. To enhance the model exposure in the networks, we incorporate a secure aggregation protocol, featuring secret sharing in federated learning~\cite{bonawitz2017practical}.

Our personalized federated learning approach requires clients to share only a portion of the model while keeping the positional embeddings (camera and image embedding) local. After the secure aggregation, the customized BEVT is rebuilt in each client by concatenating the global and local model partitions. The concatenated model is then trained on the local dataset and decoupled again for further model aggregation.

\section{Dataset}
Our federated learning approach requires a diverse dataset that comprises various vehicles equipped with different sensor installation positions and numbers, paired with BEV ground truth in different cities. However, since there is no real-world dataset that meets our requirements, we resort to using the high-fidelity simulator CARLA~\cite{dosovitskiy2017carla} and the full-stack autonomous driving simulation framework OpenCDA~\cite{xu2021opencda} to gather the necessary data. To simulate different camera sensor installation positions, we employ three distinct types of collection vehicles: a compact car, a pickup truck, and a bus. Each collection vehicle drives through various cities with consistent weather conditions and is equipped with four cameras to provide a 360-degree surrounding view, similar to~\cite{xu2022cobevt}. We employ post-processing techniques to control the number of cameras in each frame, ensuring that the collected data meets our quality standards. However, the installation poses of these cameras vary significantly, depending on the different vehicle models. In total, the car, bus, and truck datasets comprise 8352, 1796, and 1800 frames, respectively, containing 52, 14, and 9 unique scenarios. Each scenario contains diverse traffic situations and road types, following a similar collection protocol in OPV2V~\cite{xu2022opv2v}, to enrich the complexities.

\begin{figure*}[!ht]
   \centering   \includegraphics[width=1\textwidth]{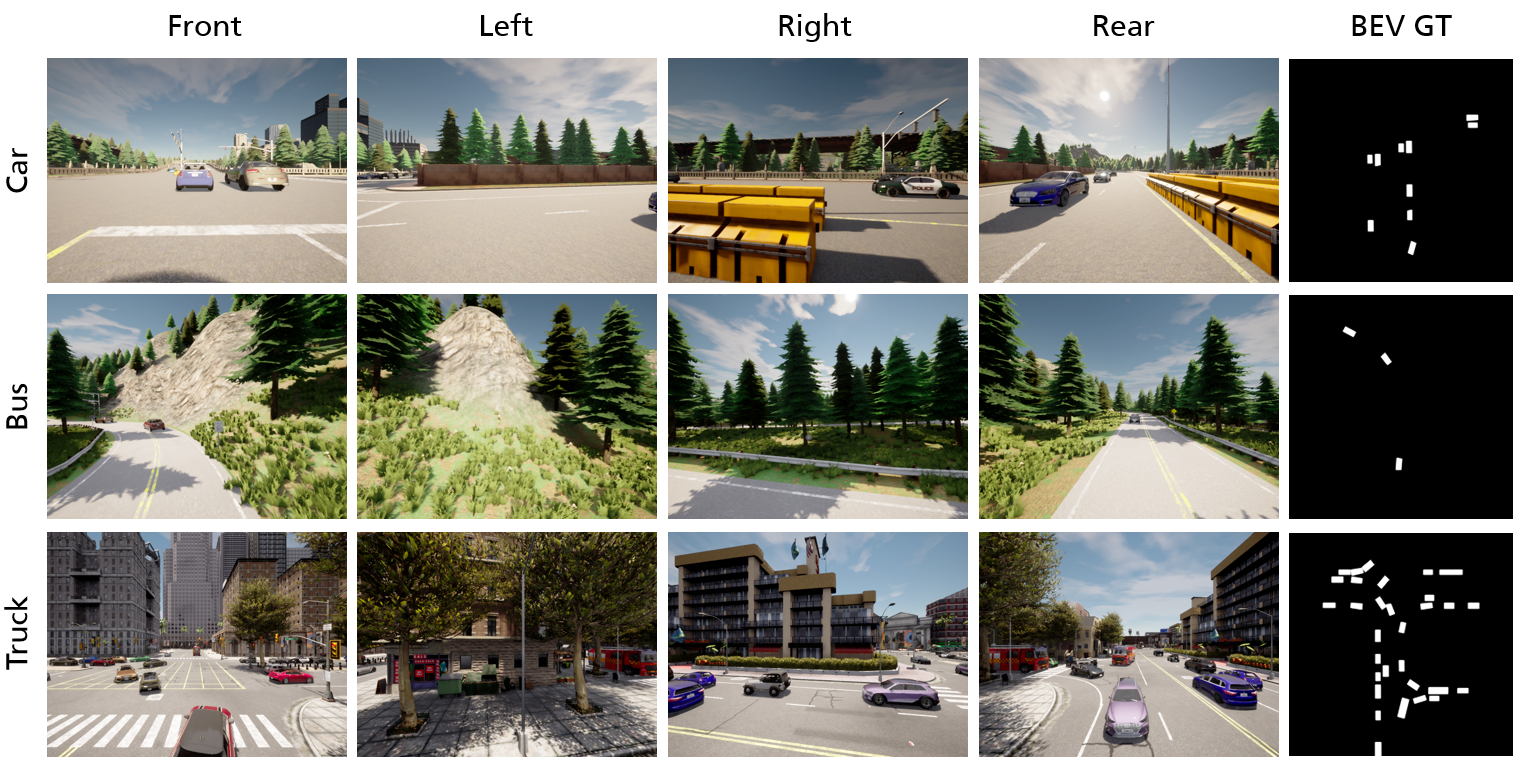}
   \caption{Visualization of four-view camera data points and BEV groundtruth (GT) for car, bus and truck clients in the FedBEVT dataset. }
   \label{fig:pose_variation}
\end{figure*}

In Tab.~\ref{table:data_sensor}, we present the sensor configuration parameters employed for data collection across a variety of vehicular clients, including cars, buses, and trucks. A qualitative illustration of the disparities in multi-camera input data from bus, truck, and car clients is provided in Figure~\ref{fig:pose_variation}.

\setlength{\tabcolsep}{3pt}
\begin{table}[ht!]
\centering
\fontsize{8}{12}\selectfont
\begin{threeparttable}
\caption{Sensor configuration parameters used for data collection in car, bus ,truck clients.}
\label{table:data_sensor}
\begin{tabular}
{C{1.0cm}|C{1.0cm}|C{1.3cm}C{1.3cm}C{1.3cm}C{1.3cm}}
\toprule 
\multicolumn{1}{c|}{Veh. Type} & Camera &  Height (m) & Roll (deg) & Pitch (deg) &  Yaw (deg)\\
\midrule
\midrule
\multirow{4}{*}{Car} & Front & 1.8 & 0 & 0 & 0 \\
& Left & 1.8 & 0 & 0 & 100\\
& Right & 1.8 & 0 & 0 & -100\\
& Rear & 1.8 & 0 & 0 & 180\\
\midrule
\multirow{4}{*}{Bus} & Front & 3.2 & 0 & -5 & 0\\
& Left & 3.2 & 0 & -5 & 100\\
& Right & 3.2 & 0 & -5 & -100\\
& Rear & 3.2 & 0 & -5 & 180\\
\midrule
\multirow{4}{*}{Truck} & Front & 4.8 & 0 & -5 & 0\\
& Left & 4.8 & 0 & -5 & 100\\
& Right & 4.8 & 0 & -5 & -100\\
& Rear & 4.8 & 0 & -5 & -80\\
\bottomrule 
\end{tabular}
\end{threeparttable}
\end{table}

\section{Experiment}
\label{sec:experiment}

\subsection{Experimental Setup}

\setlength{\tabcolsep}{3pt}
\begin{table*}[t!]
\centering
\fontsize{8}{12}\selectfont
\begin{threeparttable}
\caption{Overview of motivations and descriptions for the four use cases.}
\label{table:uc_overview}
\begin{tabular}{C{0.8cm}|L{8.5cm}|L{7.5cm}}
\toprule 
ID & Motivation & Description \\
\midrule
\midrule
UC~1 & Companies improve the model training with data from other industrial companies and open datasets. & 2 clients with data from buses and trucks, respectively; 1~additional virtual client with data from cars. \\
UC~2 & Companies improve the model training with data from each other. & 2 clients with data from buses and trucks, respectively; 2~clients with data from cars. \\
UC~3 & A model is trained on data from a number of connected vehicles, each contains only data collected in one or two specific scenarios. & 3~clients with data from buses, 4 clients with data from trucks, and 17 clients with data from cars.\\
UC~4 & Companies improve the model training with data from each other, the number of sensors at each data point is different.  & 3~clients with data from cars, the numbers of cameras on the cars from the clients are 1, 3, 4, respectively.\\
\bottomrule 
\end{tabular}
\end{threeparttable}
\end{table*}

\mypara{Use cases.} In road traffic system scenarios, the federated learning clients typically include industrial companies, such as automotive OEMs, and individual connected vehicles in public traffic. Generally, the datasets stored in these connected vehicles are relatively modest in size, while the quantity of these vehicles is substantial. In contrast, automotive corporations usually have significantly larger datasets, despite being considerably fewer in number compared to vehicles. We consider these two factors in our experimental design.

Simultaneously, each client's data originate from their respective sensor systems, resulting in data heterogeneity among clients. Our experimental design takes this source of data variance into account by adopting differing sensor system configurations across various types of vehicles (car, bus, and truck). The specific sensor system configurations used in our experiments are listed in Tab.~\ref{table:data_sensor}.

\begin{figure*}[!ht]
   \centering
   \includegraphics[width=1\textwidth]{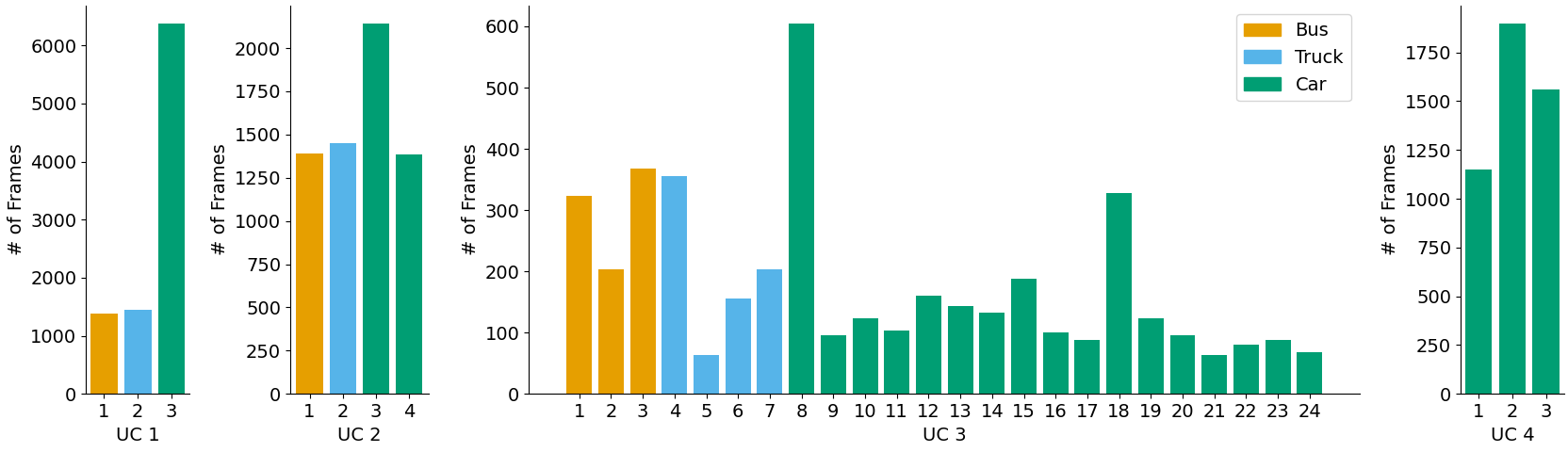}
   \caption{Distribution of federated dataset for evaluation of the FedBEVT in UC~1-4. \Crs{The specific camera configurations for each vehicle type are documented in Table~\ref{table:data_sensor}. This diversity in configuration contributes to data heterogeneity.}}
   \label{fig:data}
\end{figure*}

To comprehensively evaluate the performance of each method in potential federated learning application scenarios in road traffic systems, we define four use cases (UCs) as follows:
\begin{itemize}
    \item UC~1 considers two industrial companies, training BEVT together. The local dataset in each is collected from trucks and buses. To further enlarge the size of training dataset, an open dataset is employed in federated learning and considered as a virtual client. The dataset in this virtual client is stored in the server and set as the third data silo. 
    \item  UC~2 involves four industrial companies. The aim is to upgrade their models using the datasets from other clients, but without accessing others' raw data directly.
    \item  UC~3 represents a typical way of using public traffic data collected in vehicular networks for training, which often involves a larger number of clients (24 clients in total for federated learning). Each client owns a small amount of dataset collected from one or two specific driving scenario, e.g., a urban travel with crowded road traffic on a sunny day.
    \item UC~4 addresses federated learning for clients with different numbers of cameras, which can happen when dataset is collected from different sensor systems among manufacturers.
\end{itemize}
We summarize the motivation and the description of the four UCs in Tab.~\ref{table:uc_overview}, and the distribution of data among clients in Fig.~\ref{fig:data}. 

\mypara{BEVT architecture.}
We begin by feeding our input images $X_{i,k} \in \mathbb{R}^{L_k \times H \times W \times 3}$ through a 3-layer ResNet34 encoder.
To ensure consistency across inputs, we use the AMCM to resize all inputs to have $L'_k=4$.
The image features are then encoded at different spatial resolutions, resulting in tensors of shape $\mathbb{R}^{L'_k \times 64\times64\times128}$, $\mathbb{R}^{L'_k \times 32\times32\times256}$, $\mathbb{R}^{L'_k \times 16\times16\times512}$.
Next, we perform FAX cross attention-based transformer operations between the BEV embeddings in $\mathbb{R}^{128\times128\times128}$ (query) and the encoded image features (key and value). This step yields BEV features in $\mathbb{R}^{32\times32\times128}$.
To convert these BEV features into our final BEV results in $\{0,1\}^{256\times256}$, we use a decoder with a 3-layer bilinear upsample module.

\mypara{Hyperparameters.}
Our evaluation is conducted on a cluster with a computer cluster with 4× NVIDIA-A100-PCIE-40GB GPUs and 4× 32-Core-AMD-EPYC-7513 CPUs. The environment is a Linux system with Pytorch~1.8.1 and Cuda~11.1. During the training process, we use an initial learning rate of $2\mathrm{e}{-5}$ for a warm-up phase consisting of 20 communication rounds where the learning rate remains constant. Afterwards, we employ a Cosine Annealing learning rate scheduler~\cite{loshchilov2016sgdr}. Each client is locally trained for one epoch with a batch size of 4 using the AdamW optimizer~\cite{loshchilov2017decoupled}.

\mypara{Baselines.}
Given the limited research on federated learning on BEVT, we conduct the first trial of training fedAvg on this platform. However, to validate the effectiveness of \myMethod, we also incorporate recent research findings on federated transformer learning as a baseline on BEVT. In addition to showcasing the results of local training on each client, we compare \myMethod with the following baselines: 
\begin{itemize}
    \item FedAvg~\cite{mcmahan2017communication} is the original algorithm for federated learning, which aims to train a common global model for all clients.
    \item FedRep~\cite{collins2021exploiting} shares the data representation across clients and learns unique local heads for each client. Note that we only allows client to personalize their image encoder layers as local heads. 
    \item FedTP~\cite{li2022fedtp} uses personalized attention for each client while aggregating the other parameters among the clients. 
\end{itemize}

\subsection{Performance}

To compare the performance of \myMethod with other baselines, we compare the Average Intersection over Union (IoU) achieved by trained models in the four UCs.

\mypara{UC~1.} As shown in Tab.~\ref{table:uc1}, it is obvious that FedBEVT achieves an IoU improvement of over 50\% compared to local training, due to its indirect utilization of other clients' local training data. Additionally, \myMethod outperforms the basic FedAvg and the other two personalized federated learning approaches in the overall performance. Despite the fact that \myMethod trains a slightly more promising model to \myMethod on the bus client, the communication rounds required are significantly higher than \myMethod.

\mypara{UC~2.} Tab.~\ref{table:uc2} presents the results for UC~2, where the data volume for each client is more balanced, but still leading to outcomes similar to UC~1. In the car client~A, FedRep training shows comparable results to \myMethod, while \myMethod outperforms other methods in the other clients.

\begin{figure}[t!]
   \centering
   \includegraphics[width=0.48\textwidth]{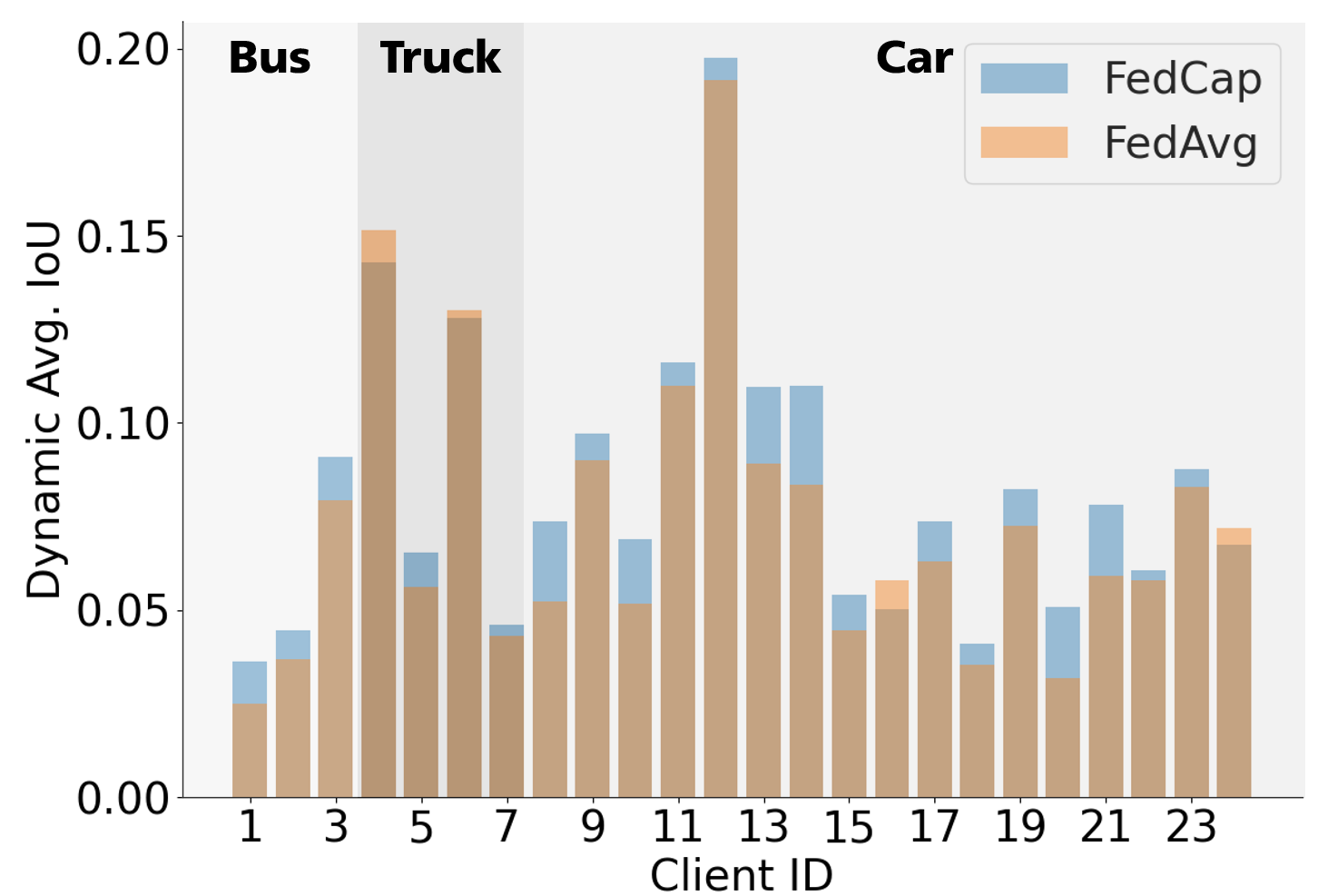}
   \caption{Comparison between \myMethod and FedAvg in UC~3, where the federated learning is organized with 24~clients including buses, trucks and cars. Each has the local data from only one or two scenarios.}
   \label{fig:uc3}
\end{figure}

\mypara{UC~3.} We conduct a targeted comparison of the performance of \myMethod and FedAvg on 24~clients with only one or two scenario data points. Since UC~3 is aimed at scenarios with poor network environments, we restrict each method to only 100~communication rounds of training. \Crs{Fig.~\ref{fig:uc3} illustrates} that more than 80\% (20 out of 24) of clients achieve superior personalized models with \myMethod.

\mypara{UC~4.}  Although AMCM  allows clients with different numbers of cameras to train a model jointly and enriches the data resources, it can also result in data heterogeneity across clients. Therefore, if the data is sufficient for training, AMCM may lead to worse training results.
As Tab.~\ref{table:uc4} shows, the client with mono-camera data can achieve better results without AMCM because the other two clients both have front cameras and can train their model based solely on that data. However, when training a model for clients with tri-cameras, AMCM can enable federated learning with more clients and improve the training performance. The performance can be further enhanced using FedCaP, which reduces the effects of AMCM.
For the client with Quad-cam, data heterogeneity becomes a bigger issue due to the significantly different data in the mono-camera client. Nonetheless, FedCaP can alleviate such effects and achieve the best model among the three approaches.

\setlength{\tabcolsep}{3pt}
\begin{table*}[t!]
\centering
\fontsize{8}{12}\selectfont
\begin{threeparttable}
\caption{Comparison between \myMethod and other baselines in UC~1.}
\label{table:uc1}

\begin{tabular}{C{2cm}|C{1.6cm}C{1.4cm}C{1.4cm}|C{1.6cm}C{1.4cm}C{1.4cm}|C{1.6cm}C{1.4cm}C{1.4cm}}
\toprule 
\multirow{3}{*}{Method} &\multicolumn{3}{c|}{Bus client} & \multicolumn{3}{c|}{Truck client} & \multicolumn{3}{c}{Virtual car client\tnote{1}}\\
 &\multicolumn{3}{c|}{$N_1=1388$ } & \multicolumn{3}{c|}{$N_2=1448$ } & \multicolumn{3}{c}{$N_3=6372$ }\\
& Train Loss & IoU$\uparrow$ & Com.$\downarrow$ &  Train Loss & IoU$\uparrow$ & Com.$\downarrow$ & Train Loss & IoU$\uparrow$ & Com.$\downarrow$\\
\midrule
\midrule
Local Training & 0.0716 &  5.42\% & - & 0.1740  & 4.16\% & - & -  & - & -\\
FedAvg~\cite{mcmahan2017communication} & 0.0343  & 16.29\%   & 315  &  0.0448 & 11.10\% & 290  &  0.0096 & 34.86\%   & 410 \\
FedRep~\cite{collins2021exploiting} & 0.0176 & \textbf{19.40}\% & 350 & 0.0303 & 14.20\%& 325 & 0.0073 & 34.37\% & 380 \\
FedTP~\cite{li2022fedtp} & 0.0281 & 10.72\% & 195 & 0.0457 & 8.82\% & 260 & 0.0122 & 33.50\% & 390 \\
\midrule
\myMethod (Ours) & 0.0528  &     \textbf{19.32\%}     & \textbf{240}  &  0.04332 &  \textbf{15.01\% }        &  295 &  0.0223 &   \textbf{35.44\%}        & 350\\
\bottomrule 
\end{tabular}
\begin{tablenotes}
     \scriptsize
     \item[1] We set the virtual car client with OPV2V~\cite{xu2022opv2v} dataset. 
   \end{tablenotes}
\end{threeparttable}
\end{table*}

\setlength{\tabcolsep}{3pt}
\begin{table*}[t!]
\centering
\fontsize{8}{12}\selectfont
\begin{threeparttable}
\caption{Comparison between \myMethod and other baselines in UC~2.}
\label{table:uc2}
\begin{tabular}{C{1.8cm}|C{1.4cm}C{1.1cm}C{0.8cm}|C{1.4cm}C{1.1cm}C{0.8cm}|C{1.4cm}C{1.1cm}C{0.8cm}|C{1.4cm}C{1.1cm}C{0.8cm}}
\toprule 
\multirow{3}{*}{Method} &\multicolumn{3}{c|}{Bus client} & \multicolumn{3}{c|}{Truck client} & \multicolumn{3}{c|}{Car client A}  & \multicolumn{3}{c}{Car client B}\\
 &\multicolumn{3}{c|}{$N_1=1388$ } & \multicolumn{3}{c|}{$N_2=1448$ } & \multicolumn{3}{c}{$N_3=2140$ }& \multicolumn{3}{c}{$N_4=1384$ }\\
& Train Loss & IoU$\uparrow$ & Com.$\downarrow$ &  Train Loss & IoU$\uparrow$ & Com.$\downarrow$ & Train Loss & IoU$\uparrow$ & Com.$\downarrow$ & Train Loss & IoU$\uparrow$ & Com.$\downarrow$\\
\midrule
\midrule
Local Training & 0.0716 & 5.42\%  & - & 0.1740 & 4.16\% & - & 0.0288  & 13.15\% & - & 0.0298  & 9.41\% & -\\
FedAvg~\cite{mcmahan2017communication} & 0.2202 & 6.92\% & 105 & 0.0483 & 5.89\% & 180 & 0.0518 & 14.09\% & 175 & 0.0899 & 14.61\% & 175\\
FedRep~\cite{collins2021exploiting} &  0.0170 & 7.45\% & 320 & 0.1319 & 7.03\% & 105 & 0.0352 &  \textbf{18.94\%} & 345 & 0.0394 & 15.41\% & 330\\
FedTP~\cite{li2022fedtp} &  0.0765 & 6.73\% & 175 & 0.1015 & 5.42\% & 140 & 0.0183 & 17.29\% & 335 & 0.0389 & 15.56\% & 300\\
\midrule
\myMethod (Ours) &  0.0061 &  \textbf{10.32\%} & 340 & 0.1294 &  \textbf{7.33\%} & 65 & 0.0162 & \textbf{18.40\%} & 360 & 0.0268 &  \textbf{16.16\%} & 370\\
\bottomrule 
\end{tabular}
\end{threeparttable}
\end{table*}

\setlength{\tabcolsep}{3pt}
\begin{table}[t!]
\centering
\fontsize{9}{12}\selectfont
\begin{threeparttable}
\caption{Evaluation of effectiveness of adaptive multi-camera masking (AMCM) in UC~4.}
\label{table:uc4}
\begin{tabular}{C{2.5cm}|C{1.5cm}C{1.5cm}C{1.5cm}}
\toprule 
\multirow{3}{*}{Method} &\multicolumn{1}{c}{Mono-cam} & \multicolumn{1}{c}{Tri-cam } & \multicolumn{1}{c}{Quad-cam}\\
& $N_1=1152$ & $N_2=1896$ & $N_3=1560$ \\
&\multicolumn{1}{c}{[1]\tnote{1}} & \multicolumn{1}{c}{[1,2,3]\tnote{1}} & \multicolumn{1}{c}{[1,2,3,4]\tnote{1}}\\
\midrule
\midrule
FedAvg\tnote{2} & \textbf{34.14\%} &  15.50\%  & 17.87\%\\
FedAvg + AMCM &  30.17\% & 19.20\% & 16.95\%  \\
\myMethod + AMCM & 31.89\% &  \textbf{20.89\%} & \textbf{20.59\%}  \\
\bottomrule 
\end{tabular}
\begin{tablenotes}
     \scriptsize
     \item[1] The numbers within the brackets represent the indices of the cameras in the perception system. Specifically, the front camera is denoted by~1, the left camera by~2, the right camera by~3, and the rear camera by~4.
     \item[2] FedAvg without AMCM: The cameras with the same setup in each client is used for federated learning. For instance, to train a model for the client of Mono-cam, all data from front camera in other clients are used for federated learning. One significant limitation of using FedAvg without AMCM is that it can only train a model for clients with a particular camera system.
   \end{tablenotes}
\end{threeparttable}
\end{table}

\subsection{Effectiveness}


To validate the personalized FedBEVT is effective in addressing the variations across local datasets in clients, \Crs{we conduct ablation experiments using UC~1 as a representative example to train models for different clients.}

We first train models using only local data and tested them on various clients. In particular, we train a model using virtual car client with OPV2V data. Although it achieves an IoU of 30.48\% on car client data, its performance significantly decreases on the bus and truck testsets (7.39\% and 2.01\%), and it even performs worse than models trained using local truck data on the truck testset. This motivates us to use FedBEVT.

Subsequently, we employ different personalized FedBEVT, namely FedTP, FedRep, and FedCaP, to train models for different clients and evaluate them on their own testsets and the testsets of other clients. The results shown in Tab.~\ref{table:AB_FedTP}, Tab.~\ref{table:AB_FedReg} and Tab.~\ref{table:AB_FedCaP} are consistent, with only locally personalized models being the most suitable for local data. Since the camera pose and quantity of a car typically do not change once it is produced, the locally collected training dataset and the data used for future inference have strong similarities. This aligns with our experimental design. Therefore, these results further emphasize the significance of personalized FedBEVT for road traffic perception in BEV.

\setlength{\tabcolsep}{3pt}
\begin{table}[t!]
\centering
\fontsize{8}{12}\selectfont
\begin{threeparttable}
\caption{Average IoU of BEV detection for each client using personalized models trained with FedTP.}
\label{table:AB_FedTP}
\begin{tabular}{C{3.5cm}|C{1.2cm}C{1.2cm}C{1.2cm}}
\toprule 
\backslashbox{Local Testset}{Per. Model}&\multicolumn{1}{c}{Bus client} & \multicolumn{1}{c}{Truck client} & \multicolumn{1}{c}{Car client}\\
\midrule
\midrule
Bus client & \textbf{10.72\%} &  4.61\%  & 9.12\% \\
\midrule
Truck client & 6.92\%  & \textbf{8.82\%} &  2.10\% \\
\midrule
Car client & 12.45\% &  6.76\% & \textbf{33.50\%} \\
\bottomrule 
\end{tabular}
\end{threeparttable}
\end{table}

\setlength{\tabcolsep}{3pt}
\begin{table}[t!]
\centering
\fontsize{8}{12}\selectfont
\begin{threeparttable}
\caption{Average IoU of BEV detection for each client using personalized models trained with FedRep.}
\label{table:AB_FedReg}
\begin{tabular}{C{3.5cm}|C{1.2cm}C{1.2cm}C{1.2cm}}
\toprule 
\backslashbox{Local Testset}{Per. Model}&\multicolumn{1}{c}{Bus client} & \multicolumn{1}{c}{Truck client} & \multicolumn{1}{c}{Car client}\\
\midrule
\midrule
Bus client & \textbf{19.40\%} &  17.73\%  & 9.39\%\\
\midrule
Truck client & 9.59\%  & \textbf{14.20\%} &  4.48\% \\
\midrule
Car client & 26.04\%  &  20.39\% & \textbf{34.37\%} \\
\bottomrule 
\end{tabular}
\end{threeparttable}
\end{table}

\setlength{\tabcolsep}{3pt}
\begin{table}[t!]
\centering
\fontsize{8}{12}\selectfont
\begin{threeparttable}
\caption{Average IoU of BEV detection for each client using personalized models trained with FedCaP.}
\label{table:AB_FedCaP}
\begin{tabular}{C{3.5cm}|C{1.2cm}C{1.2cm}C{1.2cm}}
\toprule 
\backslashbox{Local Testset}{Per. Model}&\multicolumn{1}{c}{Bus client} & \multicolumn{1}{c}{Truck client} & \multicolumn{1}{c}{Car client}\\
\midrule
\midrule
Bus client & \textbf{19.32\%} &  17.48\%  & 13.71\%\\
\midrule
Truck client & 13.01\%  & \textbf{15.01\%} &  8.89 \\
\midrule
Car client & 26.94\% & 25.26\%  & \textbf{35.44\%} \\
\bottomrule 
\end{tabular}
\end{threeparttable}
\end{table}

\subsection{Visual Analysis}

\begin{figure*}[t!]
   \centering
   \includegraphics[width=1\textwidth]{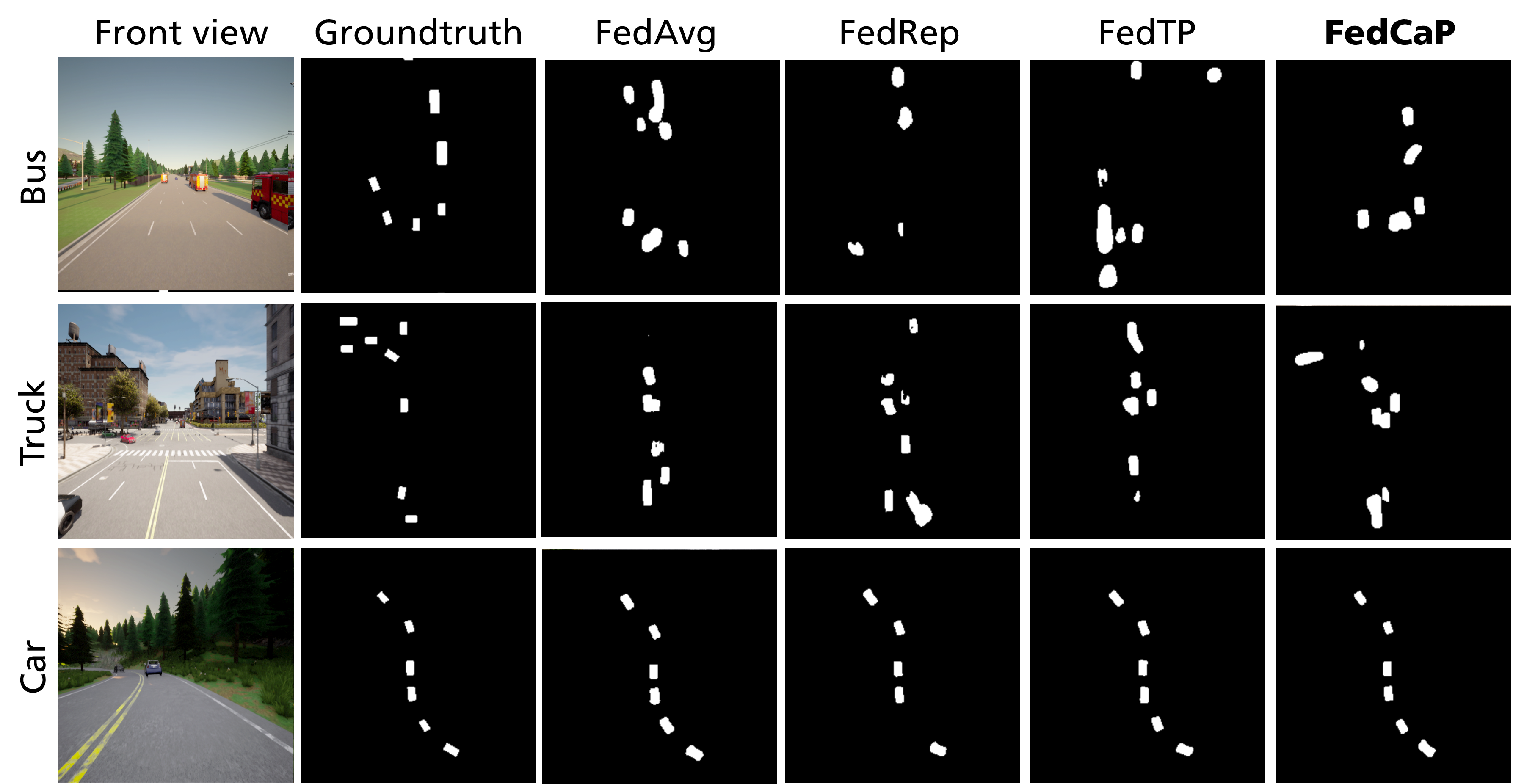}
   \caption{Visual comparison of BEV results from different federated learning approaches.}
   \label{fig:vis}
\end{figure*}

In Fig.~\ref{fig:vis}, we visually show that the BEV maps generated by \myMethod are more accurate and holistic compared to other methods in test datasets from all three vehicle types, namely bus, truck, and car. We present the front camera view of each multi-view camera system in the first column. It is evident that the camera systems vary significantly in height across different clients. Though a higher camera may offer a better BEV view, it also reduces the number of pixels occupied by each object, thereby making object detection more challenging. However, the ground truth for the BEV view remains the same -- see the second column -- as the height information in the z-axis is no longer relevant.

\mypara{FedRep}. Although FedRep achieves a relatively good ability of object recognition, it may miss some object vehicles. It trains the initial layers of the encoder to some extent to account for differences in image pixels. However, FedRep does not explicitly consider differences in the sensor perspectives. When the data heterogeneity caused by various sensor heights matters, it leads to a decrease in the object recognition ability. 

\mypara{FedTP}. We observe that the model trained by FedTP is more likely to recognize roadside trees and buildings as object vehicles. FedTP directly decouples the parameters of cross-attention, which considers data differences, but this may cause a deviation in the optimization direction of attention and the original head when the model is coupled again.

\mypara{FedCaP}. By privatizing the embeddings related to extrinsic and intrinsic parameters, FedCaP straightforwardly considers the differences in camera system configurations among clients. It results in promising overall accuracy and ability of object recognition. As shown in Fig.~\ref{fig:vis}, it is the only method that recognizes all objects compared to the other methods for the data from buses. As for the data from trucks, it is the only method that recognizes objects on the left side of a crossroad. For the data from cars, it recognizes all objects and accurately estimate their size.

\section{Conclusion}
\label{sec:conclusion}
This work investigates the efficiency of federated learning in training a transformer-based model for BEV perception in road traffic. Our analysis identifies two potential data heterogeneity issues that can impede the performance of federated learning approaches. To address these challenges, we propose two novel techniques, i.e., \myMethod and AMCM. We evaluate the effectiveness of our proposed approaches by collecting a new dataset and distributing it to clients such that typical use cases in federated settings are created. Our experimental results demonstrate that the proposed methods significantly enhance the overall performance of federated learning by personalizing the positional embeddings and increasing the data resources available for training.
In conclusion, our work highlights the potential of using federated learning for BEV perception models and presents effective solutions to overcome challenges of data heterogeneity. 

\section{Acknowledgment}
\label{sec:acknolwdgement}
The project extensively utilizes the toolchains in the OpenCDA ecosystem~\cite{10045043}, including the OpenCOOD~\cite{xu2022opv2v} and the OpenCDA simulation tools~\cite{xu2021opencda}.

\bibliographystyle{IEEEtran}
\bibliography{ref}

\vspace{11pt}

\vfill
\onecolumn
\twocolumn
\newpage
\appendices

\section{Convergence Analysis}

Based on the descriptions of FedCaP in Sec.~\ref{cap}, we first introduce the following assumptions that align with previous works in~\cite{arivazhagan2019federated, liang2020think, hanzely2021personalized, pillutla2022federated}:
\begin{assumption}{1}{L-smoothness}\label{as:1}
  The local model gradients of the $k$-th client, $\nabla_u F_k(u,v_k)$, $\nabla_v F_k(u,v_k)$, $\nabla_{uv_k} F_k(u,v_k)$ and $\nabla_{v_ku} F_k(u,v_k)$ are L-Lipschitian, i.e.,

\begin{align}
\begin{split}
||\nabla_u F_k(u^{t+1},v_k) - \nabla_u F_k(u^{t},v_k)|| & \le L_u||u^{t+1} - u^t||,\\
||\nabla_v F_k(u^{t+1},v_k) - \nabla_v F_k(u^{t},v_k)|| & \le L_{vu}||u^{t+1} - u^t||,\\
||\nabla_v F_k(u,v_k^{t+1}) - \nabla_v F_k(u,v_k^{t})|| & \le L_v||v^{t+1}_k - v_k^t||,\\
||\nabla_u F_k(u,v_k^{t+1}) - \nabla_u F_k(u,v_k^{t})|| & \le L_{uv}||v^{t+1}_k - v_k^t||,\\
\end{split}
\end{align}
and for some $\mathcal{X}$,
\begin{align}
\begin{split}
\mathcal{X} \geq \max(\frac{L_{uv}}{\sqrt{L_u L_v}}, \frac{L_{vu}}{\sqrt{L_u L_v}}).
\end{split}
\end{align}
\end{assumption}

\begin{assumption}{2}{Bounded gradient}\label{as:2}
The variance of each client's local stochastic gradient is bounded, i.e., 
\begin{align}
\begin{split}
\mathbb{E}_{B\subseteq \mathcal{D}_k}||\nabla_u F(u,v_k; B) - \nabla_u F(u,v_k)||^2 \le \sigma_u^2,\\
\mathbb{E}_{B\subseteq \mathcal{D}_k}||\nabla_v F(u,v_k; B) - \nabla_v F(u,v_k)||^2 \le \sigma_v^2.
\end{split}
\end{align}
\end{assumption}

\begin{assumption}{3}{Bounded noise}\label{as:3}
There exist $\zeta \ge 0$ and ${P} \ge 0$ applicable to all $u$ and $V$, such that,
\begin{align}
\begin{split}
||\frac{1}{K}\sum_{k=1}^k\nabla_u F(u, v_k)-\nabla_u F(u, V)||^2 \\
\le \zeta^2 + {P}^2||\nabla_u F(u,V)||^2.
\end{split}
\end{align}
In cases where $F$ does not contain a regularization term, ${P}$ equals 0.
\end{assumption}

\begin{theorem}{1}{Convergence}\label{theo:1}
FedCaP can be converged throughout the duration of the training.
\end{theorem}

\noindent\emph{Proof.} Assuming assumption~\ref{as:1} is valid, we can establish bounds for the updates of variables $u$ and $v_k$ for the k-th client, as delineated below:
\begin{align}
F_k(u^{t+1},v_k^{t+1}) - F_k(u^t, v_k^{t+1}) & \le \langle\nabla_u F_k(u^t, v_k^{t+1}),(u^{t+1}-v^t)\rangle\nonumber\\ 
& + \frac{L_u}{2} ||u^{t+1}-u^t||^2 \label{eq:8}\\
F_k(u^{t},v_k^{t+1}) - F_k(u^{t}, v_k^{t}) &\le \langle\nabla_u F_k(u^t, v_k^{t}),(u^{t+1}-v^t)\rangle \nonumber\\
& + \frac{L_v}{2} ||v_k^{t+1}-v_k^t||^2,\label{eq:9}
\end{align}
where
\begin{align}
\begin{split}
\label{eq:10}
    \langle\nabla_u F_k(u^t, v_k^{t+1}),  (u^{t+1}-v^t)\rangle &\le  \langle\nabla_u F_i(u^t, v_k^t), (u^{t+1}-u^t)\rangle \\
    & + \mathcal{X}^2 \frac{L_v}{2}||v_k^{t+1}-v_k^t||^2\\ 
    & + \mathcal{X}^2\frac{L_u}{2}||u^{t+1}-u^t||^2.
\end{split}
\end{align}
By integrating the inequalities denoted by \ref{eq:8}, \ref{eq:9}, and \ref{eq:10}, we can establish a boundary for the complete update after a single round of communication in FedCaP. This can be expressed as:
\begin{align}
\begin{split}
\label{eq:11}
&F_k(u^{t+1},v_k^{t+1}) - F_k(u^t, v_k^t) \le \langle\nabla_u F(u^t, V^t),  (u^{t+1}- u^t)\rangle\\
&+\frac{1}{K}\sum_{k=1}^K \langle \nabla_u F(u^t, v_k^t),  (v_k^{t+1} - v_k^{t})\rangle\\
&+ \frac{L_u}{2}(1+\mathcal{X}^2)||u^{t+1}-u^t||^2\\
&+ \frac{L_v}{2}(1+\mathcal{X}^2)||v_k^{t+1}-v_k^t||^2.
\end{split}
\end{align}
For inequality \ref{eq:11}, we can explore the expectation of each term:
\begin{align}
\begin{split}
\label{eq:12}
&\mathbb{E}[\langle\nabla_u F(u^t, V^t),  (u^{t+1}- u^t)\rangle] \le -\frac{E\eta_u}{2}||\nabla_u F(u^t, V^t)||^2\\
&+ \frac{\eta_u}{2}\sum_{e=0}^{E-1}\mathbb{E}||\frac{1}{K}\sum_{k=1}^K \nabla_u F(u_{k,e}^t, v_{k,e}^t) - \nabla_u F(u^t, V^t)||^2\\
&\le -\frac{E\eta_u}{2}||\nabla_u F(u^t, V^t)||^2\\
&+ \frac{\eta_u}{2K}\sum_{e=0}^{E-1}\sum_{k=1}^K\mathbb{E}||\nabla_u F(u_{k,e}^t, v_{k,e}^t) - \nabla_u F(u^t, V^t)||^2.\\
\end{split}
\end{align}
\begin{align}
\begin{split}
\label{eq:13}
&\mathbb{E}[\frac{1}{K}\sum_{k=1}^K\langle\nabla_u F(u^t, v_k^t),  (v_k^{t+1} - v_k^{t})\rangle]\\
&= -\eta_v \frac{m}{K^2}\sum_{k=1}^{K}\sum_{e=0}^{E-1}\mathbb{E}(\nabla_v F(u^t, v_k^t)  \nabla_v F(u_k^{t}- v_k^t))\\
&\le -\frac{M\eta_v e}{2K^2}\sum_{k=1}^{K}||\nabla_v F(u^t, v_k^t)||^2\\
&+ \frac{M\eta_v}{2K^2}\sum_{k=1}^{K}\sum_{e=0}^{E-1}\\
&\mathbb{E}(2\mathcal{X}^2L_uL_v||u_{k,e}^t-u^t||^2+2L_v^2||v_{k,e}^t-v_k^t||^2).\\
\end{split}
\end{align}
\begin{align}
\begin{split}
\label{eq:14}
&\mathbb{E}[\frac{L_u}{2}(1+\mathcal{X}^2)||u^{t+1}-u^t||^2] \le \frac{L_u(1+\mathcal{X}^2)\eta_u^2E^2\sigma_u^2}{2M}\\
&+ \frac{L_u(1+\mathcal{X}^2)\eta_u^2E}{2}\sum_{e=0}^{E-1}\mathbb{E}||\frac{1}{M}\sum_{k\in \mathcal{S}^t}\nabla_u F(u_{k,e}^t, v_{k,e}^t)||^2\\
&\le \frac{L_u(1+\mathcal{X}^2)\eta_u^2E^2}{2M}(\sigma_u^2+12\zeta^2(1-\frac{M}{K}))\\
&+3L_u(1+\mathcal{X}^2)\eta_u^2E^2(1+{P}^2)||\nabla F(u^t, V^t)||^2\\
&+\frac{3L_u(1+\mathcal{X}^2)\eta_uE^2}{2K}\sum_{k=1}^K\sum_{e=0}^{E-1}\\
&\mathbb{E}(L_u^2||u_{k,e}^t-u^t||^2+\mathcal{X}^2L_uL_v||v_{k,e}^t-v_k^t||^2).
\end{split}
\end{align}
\begin{align}
\begin{split}
\label{eq:15}
&\mathbb{E}[\frac{L_v}{2}(1+\mathcal{X}^2)||v_k^{t+1}-v_k^t||^2] \\
&\le \frac{L_v(1+\mathcal{X}^2)M}{2K^2} \sum_{k=1}^K\mathbb{E}(||v_{k,E}^t-v_k^t||^2)\\
&\le \frac{3L_v(1+\mathcal{X}^2)\eta_v^2 E^2M}{2K^2} \sum_{k=1}^K(||\nabla_v F(u^t,v_k^t)||^2 \\
&+ \sum_{e=0}{E-1}\mathbb{E}(L_v^2||v_{k,e}^t-v_k^t||^2 + \mathcal{X}^2L_uL_v||u_{k,e}^t-u^t||^2))\\
&+ \frac{L_v(1+\mathcal{X}^2)\eta_v^2E^2M\sigma_v^2}{2K}.
\end{split}
\end{align}
By integrating \ref{eq:13}, \ref{eq:14}, \ref{eq:15} and \ref{eq:16} into \ref{eq:12}, we obtain the following inequality for the overall update of a communication round in FedCaP:
\begin{align}
\begin{split}
\label{eq:16}
&\mathbb{E} [F(u^{t+1}, V^{t+1}) - F(u^{t}, V^{t})]  \le -\frac{\eta_u * E}{4}||\nabla_u F(u^t, V^t)||^2 \\
& - \frac{\eta_vEM}{4K^2} \sum_{k=1}^K||\nabla_v F(u^t, v_k^t)||^2 \\
& + \frac{L_u(1+\mathcal{X}^2)\eta_u^2E^2}{2}(\sigma_u^2 + \frac{12\zeta^2}{M}(1-M/K))\\
& + \frac{L_v(1+\mathcal{X}^2)\eta_v^2 E^2 \sigma_v^2 M}{2K} \\
& + \frac{2}{K}\sum_{k=1}^K\sum_{e=0}^{E-1}\mathbb{E}||u_{k,e}^t - u^t||^2 (L_u^2 \eta_u + \frac{M}{K}\mathcal{X}^2L_uL_v\eta_v)\\
& +\frac{2}{K}\sum_{k=1}^K\sum_{e=0}^{E-1}\mathbb{E}||v_{k,e}^t - v_k^t||^2(\frac{M}{K}L_v^2\eta_v + \mathcal{X}^2 L_u L_v \eta_u).
\end{split}
\end{align}
Next, we evaluate the boundaries of the last two terms in inequality \ref{eq:16}. Assuming that assumptions~\ref{as:2} and \ref{as:3} are valid, and after simplifying the constants within the terms, we obtain the following:
\begin{align}
\begin{split}
&\mathbb{E} [F(u^{t+1}, V^{t+1}) - F(u^{t}, V^{t})]  \le - \frac{C||\nabla_u F(u^t, V^t)||^2}{8L_u} \\
&- \frac{CM\frac{1}{K}\sum_{k=1}^K||\nabla_v F(u^t, v_k^t)||^2}{8L_vK} \\
&+ C^2(1+\mathcal{X}^2) (\frac{\sigma_u^2}{2L_u} + \frac{M \sigma_v^2}{nL_v} +  \frac{6\zeta^2}{L_vM}(1-\frac{M}{K})) \\
&+ C^3(1+\mathcal{X}^2)(1-\frac{1}{E})(\frac{24\zeta^2}{L_u} + \frac{4\sigma_u^2}{L_u} + \frac{4\sigma_v^2}{L_u}),
\end{split}
\end{align}
with learning rate $\eta_u \le (12EL_u(1+\mathcal{X}^2)(1+{P}^2))^{-1}$ and $\eta_v \le (6EL_v(1+\mathcal{X}^2))^{-1}$, and $C>0$ represents some positive constant.
Finally, the convergence rate of FedCaP can be expressed as follows:
\begin{align}
\begin{split}
&\frac{1}{T}\sum_{t=0}^{T-1}(\frac{1}{L_u} \mathbb{E}||\nabla_u F(u^t, V^t)||^2 + \frac{M}{L_v K^2}\sum_{k=1}^K \mathbb{E}||\nabla_v F(u^t, v^t_k)||^2) \\
&\le \sqrt[]{\frac{(F(u^0,V^0)-F^*)(1+\mathcal{X}^2)(\frac{\zeta^2}{L_u}(1-\frac{M}{K})+\frac{\sigma_u^2}{L_u}+\frac{M\sigma_v^2}{KL_v})}{T}} \\
&+ \sqrt[3]{\frac{(F(u^0,V^0)-F^*)^2 (1+\mathcal{X}^2)(\frac{\zeta^2}{L_u}+\frac{\sigma_u^2}{L_u}+\frac{\sigma_v^2}{L_v})(1-\frac{1}{E})}{T^2}}\\
&+ \frac{(F(u^0,V^0)-F^*)(1+\mathcal{X}^2)(1+{P}^2)}{T}\\
&+ \frac{(F(u^0,V^0)-F^*) \sqrt[]{\frac{K}{M}(1-\frac{1}{E})(1+\mathcal{X}^2)(1+{P}^2)}}{T}\\
&\le  \mathcal{O}(\frac{1}{\sqrt{T}}) + \mathcal{O}(\frac{1}{\sqrt[3]{T^2}}) + \mathcal{O}(\frac{1}{T}),
\end{split}
\end{align}
$F(u^0,V^0)$ represents the function $F$ with initial values of $u^0$ and $V^0$, while $F^*$ encompasses all other instances of $F$. It is evident that the convergence rate of FedCaP approximates $1/\sqrt{T}$, aligning with the vanilla Stochastic Gradient Descent (SGD), but this holds true only under the following conditions:
\begin{align}
\begin{split}
\label{condition1}
T \geq \frac{KL_vL_u(1+{P}^2)(F(u^0,V^0)-F^*)}{\zeta^2L_v(K-M)+\sigma_u^2L_v+ML_u\sigma_v^2}.
\end{split}
\end{align}
In the scenario where all clients are selected in each round and there are no regularization terms in the local loss functions, we can reformulate condition \ref{condition1} as follows:
\begin{align}
\begin{split}
\label{condition2}
T \geq \frac{KL_vL_u(F(u^0,V^0)-F^*)}{\sigma_u^2L_v+KL_u\sigma_v^2}.
\end{split}
\end{align}

\end{document}